\documentclass[conference]{IEEEtran}
\IEEEoverridecommandlockouts
\usepackage{cite}
\usepackage{amsmath,amssymb,amsfonts}
\usepackage{algorithmic}
\usepackage{graphicx}
\usepackage{textcomp}
\usepackage{xcolor}
\usepackage[font=small]{caption}
\usepackage{subcaption}
\usepackage{romannum}
\usepackage{booktabs}
\usepackage{subfig}
\usepackage{dblfloatfix}
\begin{document}



\title{\fontsize{21}{28}\selectfont Knowledge Distillation from Language-Oriented to Emergent Communication for Multi-Agent Remote Control}

\author{\IEEEauthorblockN{Yongjun Kim, $^*$Sejin Seo, $^\dagger$Jihong Park, $^\ddagger$Mehdi Bennis, $^*$Seong-Lyun Kim, and Junil Choi}
\thanks{Y. Kim and J. Choi are with the School of Electrical Engineering, KAIST, Daejeon 34141, Korea (email: \{yongjunkim, junil\}@kaist.ac.kr).}
\thanks{$^\dagger$J. Park is with the School of Information Technology, Deakin University, VIC 3218, Austraila (email: jihong.park@deakin.edu.au).} 
\thanks{$^\ddagger$M. Bennis is with the Centre for Wireless Communications, University of Oulu, Oulu 90570, Finland (email: mehdi.bennis@oulu.fi).}
\thanks{$^*$S. Seo and S.-L. Kim are with the School of Electrical \& Electronic Engineering, Yonsei University, Seoul 03722, Korea (email: \{sjseo, slkim\}@ramo.yonsei.ac.kr).} 
\thanks{J. Choi and J. Park are corresponding authors.}
}

\maketitle

\begin{abstract}
In this work, we compare emergent communication (EC) built upon multi-agent deep reinforcement learning (MADRL) and language-oriented semantic communication (LSC) empowered by a pre-trained large language model (LLM) using human language. In a multi-agent remote navigation task, with multimodal input data comprising location and channel maps, it is shown that EC incurs high training cost and struggles when using multimodal data, whereas LSC yields high inference computing cost due to the LLM's large size. To address their respective bottlenecks, we propose a novel framework of language-guided EC (LEC) by guiding the EC training using LSC via knowledge distillation (KD). Simulations corroborate that LEC achieves faster travel time while avoiding areas with poor channel conditions, as well as speeding up the MADRL training convergence by up to 61.8\% compared to EC.
\end{abstract}


\section{Introduction}

Semantic communication (SC) is an emerging research paradigm aimed at designing efficient and effective communication for specific tasks \cite{Gunduz:2023}. Unlike traditional communication systems centered on the accurate delivery of raw data, SC focuses on conveying effective meanings or intention for given tasks, also referred to as semantic representations (SRs). Machine learning (ML) plays a crucial role in extracting these SRs from complex raw data, and recent works in this area can be broadly categorized into bottom-up and top-down approaches. 
\begin{figure}[t]
    \centering
    \includegraphics[scale=0.28]{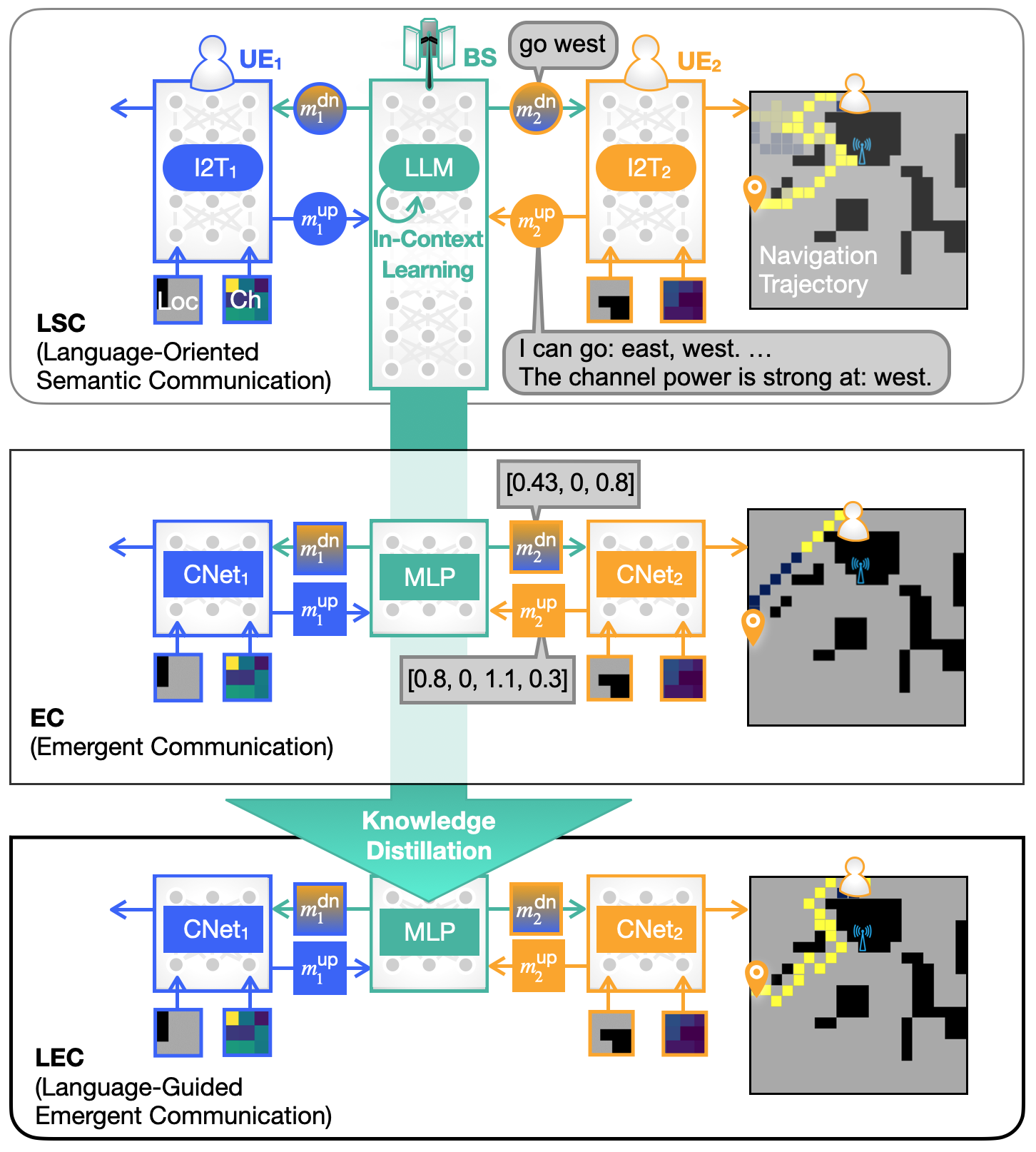}
    \caption{ A schematic illustration of: LSC (top), EC (middle), and LEC (bottom).}
    \label{fig:summary}
\end{figure}
The bottom-up approach allows SRs to naturally emerge from multi-way task-specific interactions. A prominent example is deep joint source and channel coding (DeepJSCC) for data reconstruction or other inference tasks \cite{Bourtsoulatze:2019}, wherein SRs emerge as latent variables within an encoder-decoder architected neural network (NN). Another example is emergent communication (EC) for multi-agent control tasks \cite{Foerster:2016,seo2023towards}. In EC, NN-structured agents exchange their hidden-layer activations by means of a multi-agent deep reinforcement learning (MADRL) framework. These activations, which are initially random values, become meaningful SRs for inter-agent coordination after training completes. While provisioning fast inference speed after training, the MADRL training process is costly and prone to bias, particularly when dealing with multi-modal data.

Conversely, the top-down approach first imposes a pre-defined structure on SRs, and transforms raw data into such structured SRs. Language-oriented SC (LSC) falls into this category, which considers natural language as a pre-defined structure to leverage the capabilities of large language models (LLMs) built upon natural language \cite{Touvron:2023,Li:2022,Nam:2023,Jie:2022,Hao:2023}. Precisely, LSC first utilizes pre-trained multi-modal generative models to convert raw data samples into text embeddings, such as the BLIP model for image-to-text conversion \cite{Li:2022}. These text-based SRs instruct a pre-trained LLM to carry out various tasks including compression \cite{Nam:2023}, context reasoning \cite{Jie:2022}, and control decision-making \cite{Hao:2023}. While LLMs consume a large computing cost per inference, they can instantly adapt to new tasks by providing demonstration with examples. This unique capability of LLMs is called in-context learning, which bears a significant advantage over EC that requires MADRL re-training for such adaptation.

To harness the merits of both approaches, we propose a novel framework of language-guided EC (LEC) by integrating LSC into EC. As illustrated in Fig.~\ref{fig:summary}, LEC employs an in-context learned LSC as a teacher to guide an EC as its student via knowledge distillation (KD), an ML technique to synchronize outputs for common inputs \cite{Hinton:2014}. To demonstrate the feasibility of LEC, we consider a multi-agent remote control and navigation task, wherein agents aim to reach their target destinations while avoiding poor channel conditions. During the task, each agent perceives its local location and channel maps while exchanging SRs with a base station (BS). 

We verify that EC is biased towards location maps, leading to frequent encounters with poor channels. LSC avoids poor channels, but the trajectories vary significantly over episodes. This variability is presumably due to the LLM's enforcing inference despite uncertainty, also known as LLM hallucinations, and resultant error propagation in the LLM's memory or within its context window. To address this, we develop a method to refine and select the top-$L$ LSC trajectories for teacher knowledge, and a modified KD method to reflect this. Simulations show that LEC achieves $61.8$\% less average training steps to convergence compared to EC. Furthermore, LEC achieves low computing costs during both training and inference, thanks to in-context learning of LSC and training convergence acceleration of KD in EC.

\section{System model}

\begin{figure}[t]
  \centering
  \begin{subfigure}{.23\columnwidth}
    \includegraphics[width=\linewidth]{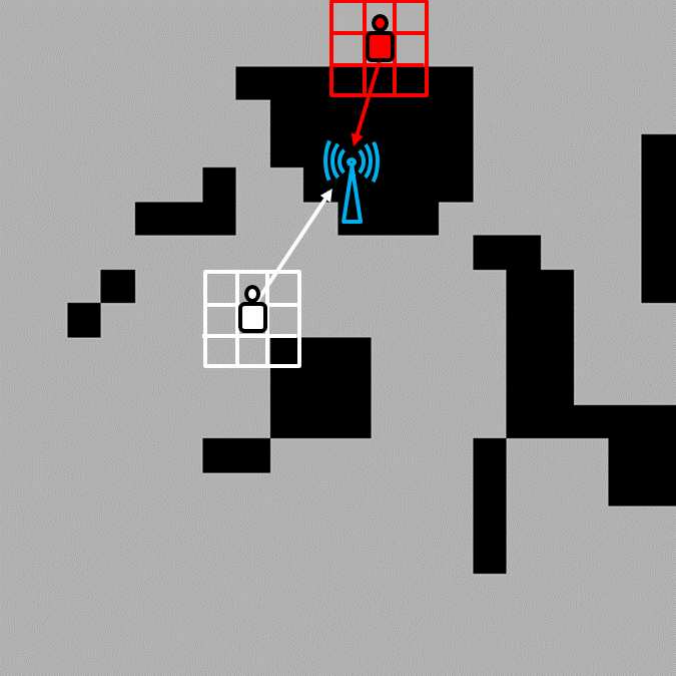}
    \caption{Location map.}
    \label{global location map}
  \end{subfigure}\hfill
  \begin{subfigure}{.26\columnwidth}
    \includegraphics[width=\linewidth]{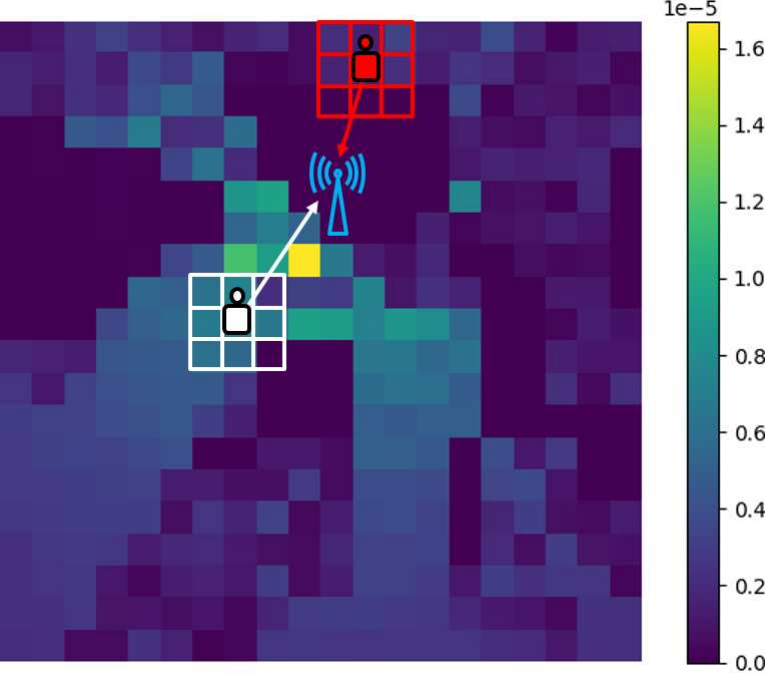}
    \caption{Channel map.}
    \label{global channel map}
  \end{subfigure}\hfill
  \begin{subfigure}{.23\columnwidth}
    \includegraphics[width=\linewidth]{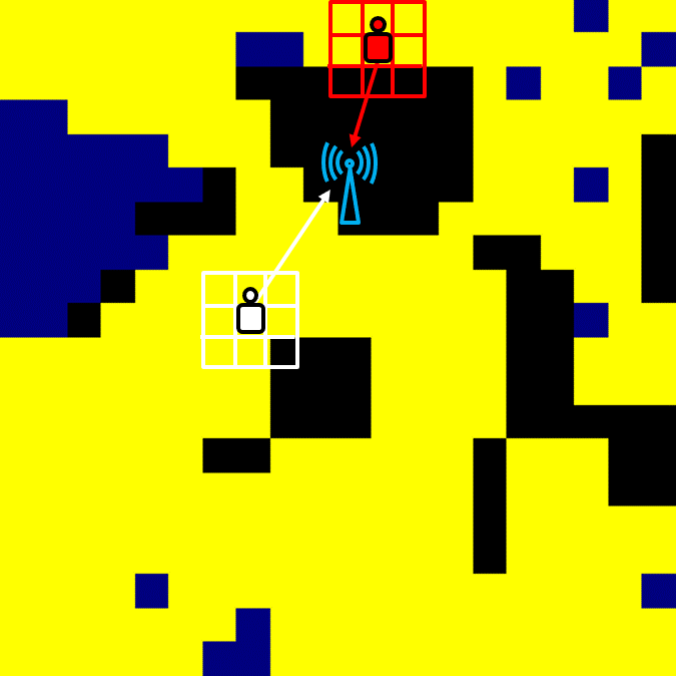}
    \caption{Channel map\\ (2-level with\\ $\sqrt{\eta} = 3 \times 10^{-7}$).}
    \vspace{-0.78cm}
    \label{003}
  \end{subfigure}\hfill
  \begin{subfigure}{.23\columnwidth}
    \includegraphics[width=\linewidth]{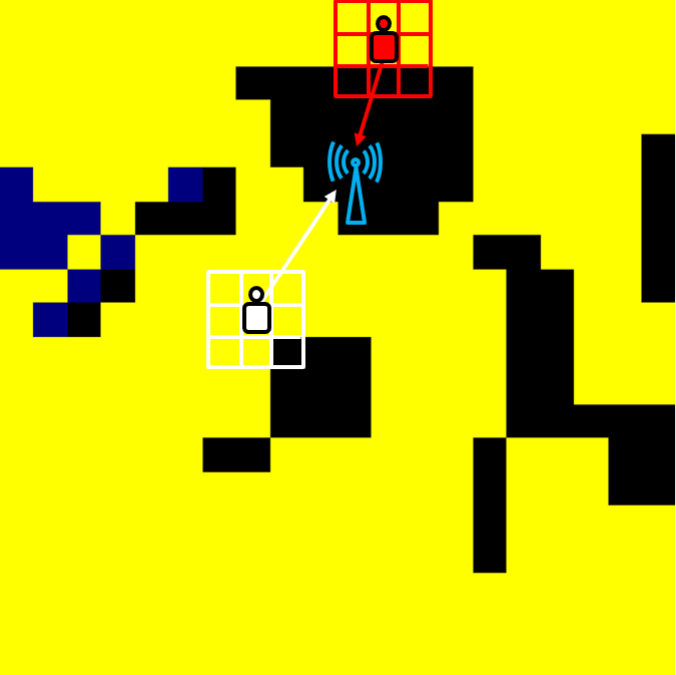}
    \caption{Channel~map\\ (2-level with\\ $\sqrt{\eta} = 10^{-45}$).}
    \vspace{-0.78cm}
    \label{45}
  \end{subfigure}
  \vspace{0.7cm}
  \caption{Global map wherein each UE observes within $3\times 3$ FoV.}
  \label{global map}
\end{figure}

\subsection{Network and Channel Model} 
The network under study consists of a single BS associated with a set $\mathcal{J}$ of user equipments (UEs). In a square grid $\mathcal{M}\subset \mathbb{R}^2$, the location $\mathbf{z}_\text{bs}\in \mathcal{M}$ of the BS is fixed at the top of a building, as shown by the global location map in Fig.~\ref{global map}(a) where the area $\mathcal{M}_b\subset \mathcal{M}$ of buildings is colored in black. At a time step $t$, the $j$-th UE is located at $\mathbf{z}_j^t \in \mathcal{M}\backslash\mathcal{M}_b$. To coordinate and control multiple agents, the BS broadcast a downlink (DL) message $m^{\text{dn}, t}$ with power $P_j^{\text{dn},t}$ to all UEs. Assuming sufficiently large $P_j^{\text{dn},t}$, we consider that DL communication is error-free, and the received DL message at every UE is $m^{\text{dn}, t}$. 

By contrast, each UE has a limited transmit power $P_j^{\text{up},t}$, leading to erroneous uplink (UL) message reception under noisy channels. Precisely, to report local information to the BS, the $j$-th UE transmits an UL message $m_j^{\text{up},t}$, and the received UL message at the BS is:
\begin{align}
\tilde{m}_j^{\text{up},t} = h_{\mathbf{z}_j^t}\sqrt{P_j^{\text{up},t}}m_j^{\text{up},t} + n,
\label{eq:ul}
\end{align}
where $h_{\mathbf{z}_j^t}$ is the UL wireless channel from $\mathbf{z}_j^t$ to $\mathbf{z}_\text{bs}$, and $n \sim \mathcal{CN}(0, \sigma^2)$. 
According to \eqref{eq:ul}, the received UL signal-to-noise ratio (SNR) is $P_j^{\text{up},t} |h_{\mathbf{z}_j^t}|^2 / \sigma^2$. Assuming time division duplexing (TDD), the UL channel gain $|h_{\mathbf{z}_j^t}|^2$ is reciprocal with its DL channel gain, which is visualized by the global channel map in Fig.~\ref{global map}(b). To maintain a constant target received SNR $P_r / \sigma^2$ at the BS, the UE applies channel inversion power control such that the transmit power $P_j^{\text{up},t}$ is set as $P_r/ |h_{\mathbf{z}_j^t}|^2$.









\subsection{Multi-Agent Remote Navigation Task} \label{Sec:Task}
The $j$-th UE is initially located at $\mathbf{z}_j^0$, and moves to reach a pre-determined destination ${\mathbf{\hat{z}}}_{j}\in \mathcal{M}\backslash\mathcal{M}_b$. Each UE aims to achieve two objectives: 1) minimizing the number of $T_j$ time stamps required to reach its destination, such that $\mathbf{z}_j^{T_j} = {\mathbf{\hat{z}}}_{j}$ for all $j \in \mathcal{J}$; while 2) avoiding the locations in $\mathcal{M}_w\subset \mathcal{M}\backslash\mathcal{M}_b$ associated with weak channels. Given the channel inversion power control, the weak channel condition is defined as $ |h_{\mathbf{z}_j^t}|^2 < \eta$ with $\eta:=P_r/P_{\text{th}}$ in which the UE transmit power exceeds a target power budget $P_{\text{th}}$, i.e., $P_j^{\text{up},t} > P_{\text{th}}$.



To achieve this goal, at each time $t$, the $j$-th UE observes its local location map $L_j^t$ and local channel map $C_j^t$ within its $3 \times 3$ field of view (FoV), as depicted in Fig.~\ref{global map}. To report this multimodal data $\{L_{j}^t, C_{j}^t\}$, the UE transmits an UL message $m_j^{\text{up},t}$ to the BS. The BS receives the UL message collection $\tilde{m}^{\text{up},t} = \{\tilde{m}_1^{\text{up}, t},\tilde{m}_2^{\text{up}, t},\ldots,\tilde{m}_{\lvert \mathcal{J} \rvert}^{\text{up}, t}\}$ from all UEs, and thereby generates a DL message $m^{\text{dn}, t}$ that is broadcasted to all UEs. Based on the received $m^{\text{dn}, t}$, the $j$-th UE takes an action $a_j^t$ from an action space $ \mathcal{A}  := \{(-v ,\! -v ),\!(-v , 0),\!(-v , v ),\!(0 , \!-v ),\! (0 , v ),\!(v ,\! -v ),\! (v , 0),\!(v , v ) \}$, 
where $v$ is the unit traveling distance in one time step. There are invalid actions that result in inter-UE collision, out-of-area, and building blockage, i.e., $\mathbf{z}_j^{t+1} = \mathbf{z}_i^{t+1}$ with $i\neq j$, $\mathbf{z}_j^{t+1}\notin \mathcal{M}$, and $\mathbf{z}_j^{t+1} \in \mathcal{M}_b$, respectively. For such invalid actions, the UE remains the previous position, i.e., $\mathbf{z}_j^{t+1} = \mathbf{z}_j^t $.




\begin{figure*}[!b]
\hrulefill
\footnotesize

\begin{align}\setcounter{equation}{3}
\theta_\text{EC}^\star &= \mathop{\mathrm{arg\,min}}_{\theta} \Bigg\{ \mathbb{E} \Bigg[ \sum_{j,t} \Big[\{r_j^t + \gamma \max_{{a'}_j^{t}} Q'({s'}_j^{t
},{a'}_j^{t}|\theta) 
 - Q(s_j^t,a_j^t|\theta)\} ^2\Big] \Bigg] \Bigg\} 
 \label{EC loss function}
 \end{align}
 \begin{align}
 \setcounter{equation}{11}
 \theta_\text{LEC}^{\star} &= \mathop{\mathrm{arg\,min}}_{\theta} \Bigg\{\mathbb{E} \Bigg[ \sum_{j,t}  \Big[\{\tilde{r}_j^t + \gamma \max_{{a'}_j^{t}} Q'({s'}_j^{t
},{a'}_j^{t}|\theta) 
 - Q(s_j^t,a_j^t|\theta)\} ^2 + \lambda \mathcal{L}_\text{KLD} \Big] \Bigg] \Bigg\}
 \label{LEC loss function}
\end{align}

\end{figure*}

\section{Emergent Communication versus Language-Oriented Semantic Communication} \label{Sec:ECvsLSC}
In this section, we introduce EC and LSC. Throughout the paper, EC exchanges modulated symbols optimized during MADRL training, while LSC exchanges natural language messages digitally encoded using $8$-bits ASC\Romannum{2} code and modulated by $16$QAM.

\subsection{Emergent Communication}

The $j$-th UE has at time slot $t$ has an input state $s_j^t = \{s_{j_1}^t, s_{j_2}^t\}$, where $s_{j_1}^t:= \{L_j^t, C_j^t \}$ involves a local location map and a local channel map while $s_{j_2}^t$ is the image of the UE's movement trajectory until $t$. For EC framework depicted in the middle of Fig.~\ref{fig:summary}, a model $G_j (\cdot)$ named CNet generates the current action $a_j^t$, subsequent hidden state of a recurrent neural network (RNN) $h_j^{t+1}$, and the UL message $m_j^{\text{up},t}$ from the current state $s_j^t$ and the last received DL message ${m}_j^{\text{dn},t}$, i.e., $(a_j^t , h_j^{t+1}, m_j^{\text{up},t}) = G_{j}(s_j^t , h_j^t , {m}_j^{\text{dn},t}).$ The BS employs multilayer perceptron (MLP) $G_{\text{bs}}(\cdot)$ to generate DL messages from the UL messages as  $m^{\text{dn},t+1} = G_{\text{bs}}(\tilde{m}^{\text{up},t})$. 
The set of the parameters of $G_{\text{bs}}(\cdot)$ and $G_j(\cdot)$ denoted as $\theta := \{\theta_\text{bs},\cup_{j=1}^{|\mathcal{J}|} \theta_j\}$ is optimized by minimizing the Deep-Q learning loss function as (\ref{EC loss function}). Here, $Q(\cdot)$ and $Q'(\cdot)$ represent the current and target $Q$ networks, respectively. The state-action pairs $(s_j^t,a_j^t)$ and $({s'}_j^t,{a'}_j^t)$ denote the inputs to the respective Q-networks. The reward function $r_j^t$ is defined as:


\begin{equation}\tag{2}
    r_{j}^t = 10 \cdot \mathbf{1}_{\mathbf{z}_j^t = \mathbf{\hat{z}}_{j}} - 0.1 \cdot \mathbf{1}_{\mathbf{z}_j^{t+1} = \mathbf{z}_j^t} -0.1 \cdot \mathbf{1}_{|h_{\mathbf{z}_j^t}|^2 \leq \eta} -0.01 \cdot \mathbf{1}_{\mathbf{z}_j^t \ne \mathbf{\hat{z}}_{j}},
    \label{reward function}
\end{equation}
where the indicator function  $\mathbf{1}_{\text{A}}$ becomes one if A is true and otherwise we have zero. In  (\ref{reward function}), each condition describes (\romannum{1}) reaching the target, (\romannum{2}) invalid action, (\romannum{3}) exceeding the power budget,  and (\romannum{4}) time consumption, respectively. Based on the acquired $\theta_{\text{EC}}^{\star}$, the policy is determined by
\begin{align}\tag{3}
\text{(\textbf{EC})}\quad a_j^t = \mathop{\mathrm{arg\,max}}_a p(a | s_j^t, h_j^t, \tilde{m}_j^{\text{dn},t} ; \theta_{\text{EC}}^{\star}),\label{Eq:EC}
\end{align}
where $\theta_{\text{EC}}^{\star}$ denotes the trained model parameters given in \eqref{EC loss function}.

\subsection{Language-Oriented Semantic Communication}

In LSC, as illustrated in Fig.~\ref{fig:summary}, each UE employs an image-to-text (I2T) generative model $G_{\text{Gen},j}(\cdot)$ to transform $s_{j_1}^t$ into a natural language message $m_{j_1}^{\text{up},t} = G_{\text{gen},j}(s_{j_1}^t)$. The message $m_{j_1}^{\text{up},t}$ is complemented by additional textual information $m_{j_2}^{\text{up},t}$ that provides direction guidance towards the destination $\mathbf{\hat{z}}_{j}$ by reporting the remaining taxicab distance to the destination. Consequently, the UL message is given by $m_j^{\text{up},t} = \{m_{j_1}^{\text{up},t}, m_{j_2}^{\text{up},t}\}$. The BS aggregates the received UL messages from all UEs, i.e., $\tilde{m}^{\text{up},t} = \{\tilde{m}_1^{\text{up},t}, \tilde{m}_2^{\text{up},t}, \ldots, \tilde{m}_{|\mathcal{J}|}^{\text{up},t}\}$, and produces the DL message $m^{\text{dn},t} = G_{\text{LLM}}(\tilde{m}^{\text{up},t}) = \{m^{\text{res},t}, m_1^{\text{a},t},m_2^{\text{a},t}, \ldots, m_{|\mathcal{J}|}^{\text{a},t}\}$ using an LLM  $G_{\text{LLM}}(\cdot)$. The DL messages involve the message $m_j^{\text{a},t}$ to instruct an action to the $j$-th UE, as well as the message $m^{\text{res},t}$ explaining the rationale behind the LLM's action decisions.

The current LLM often struggles when dealing with continuous values \cite{Touvron:2023,Li:2022,Nam:2023,Jie:2022,Hao:2023}. To detour this, we quantize the amplitudes of the original global channel map in Fig.~\ref{global channel map} into two levels, before inputting $s_{j_1}^t$ into $G_{\text{gen},j}(\cdot)$ and $G_{j}(\cdot)$. Such quantization aligns with the goal of our task, avoiding weak channel conditions associated with the areas $\mathcal{M}_w$, as described in Sec.~\ref{Sec:Task}. As visualized by the quantized global channel maps in Fig.~\ref{003} and \ref{45}, dark blue regions represent $\mathcal{M}_w$, which decrease with the transmit power budget $P_\text{th}$.


In EC, task-specific knowledge is learned and stored in $\theta_\text{EC}^\star$, and therefore actions in \eqref{Eq:EC} are determined by only providing current observations and received messages. In LSC, by contrast, the pre-trained LLM $\theta_\text{LLM}$ is general knowledge, which becomes task-specific by additionally providing task-related text prompts. Such additional prompts include: the BS's history of previously received UL messages  and produced DL messages ${m}^\text{bs}_{t-1}= \cup_{u=1}^{t-1} \{\tilde{m}^{\text{up},u},m^{\text{dn},u}\}$, a meta instruction $x'$ outlining task outlining task guidelines and the role of the BS, and $K$-pair examples $c_K = \{(i_1, o_1), (i_2, o_2), \ldots, (i_K, o_K)\}$. Consequently, the action $a_j^t$ in LSC are determined as follows:
\begin{align}\tag{5}
\text{(\textbf{LSC})}\quad a_j^t = F\big(\mathop{\mathrm{arg\,max}}_m p(m | \tilde{m}^{\text{up},t} , {m}^{\text{bs}}_{t-1},  c_K;\theta_{\text{LLM}})\big),
\label{llm action}
\end{align}
Here, the function $F(\cdot)$ converts a text message $m_i\in\mathcal{A}_m$
into an action $a_i\in\mathcal{A}$, i.e., $a_i = F(m_i)$, where $\mathcal{A}_m:=\{\text{`northwest'}, \text{`north'},\text{`northeast'}, \text{`west'},\text{`east'},\text{`southwest'},\\\text{`south'}, \text{`southeast'} \}$.





\begin{figure}[t]
  \centering
  \begin{subfigure}[b]{\columnwidth}
    \includegraphics[width=\columnwidth]{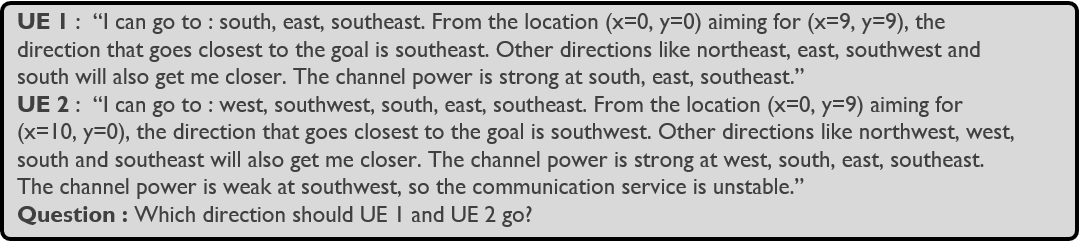} \vspace{-0.5cm}
    \caption{UL messages $\tilde{m}^{\text{up},t}$ at SNR = 20dB.}
    \label{ul 20db}
    \vspace{0.1cm}
  \end{subfigure}
  \begin{subfigure}[b]{\columnwidth}
    \includegraphics[width=\columnwidth]{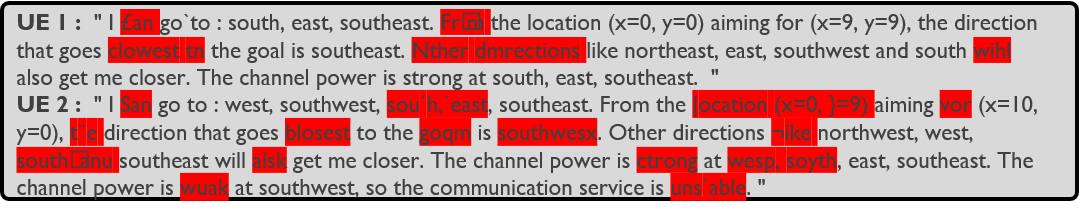} \vspace{-0.6cm}
    \caption{UL messages $\tilde{m}^{\text{up},t}$ at SNR = 15dB.}
    \label{ul 15db}
    \vspace{0.1cm}
  \end{subfigure}
  \begin{subfigure}[b]{0.32\columnwidth}
    \includegraphics[width=1\linewidth]{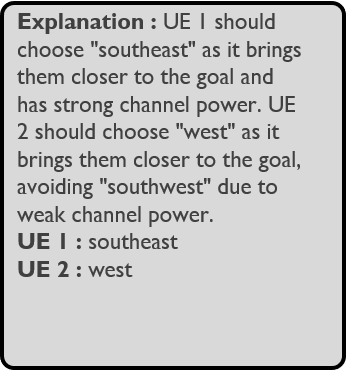}
    \caption{DL message ${m}^{\text{dn},t}$ at SNR = 20dB, $K=2$.}
    \label{dl 20db 2shot}
  \end{subfigure}
   \begin{subfigure}[b]{0.315\columnwidth}
    \includegraphics[width=\linewidth]{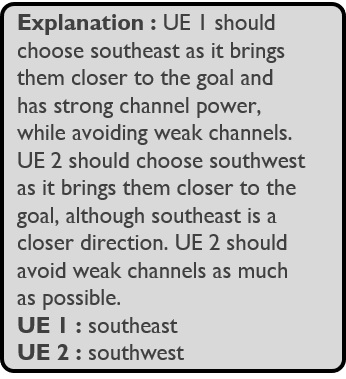} 
    \caption{DL message ${m}^{\text{dn},t}$ at SNR = 15dB, $K=2$.}
    \label{dl 15db 2shot}
  \end{subfigure}
    \begin{subfigure}[b]{0.318\columnwidth}
    \includegraphics[width=\linewidth]{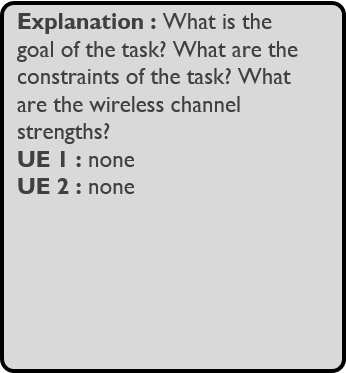} 
    \caption{DL message ${m}^{\text{dn},t}$ at SNR = 15dB, $K=0$.}
    \label{dl 15db 0shot}
  \end{subfigure}
  \caption{Example prompts of LSC.}
  \label{LSC_prompts}
\end{figure}

In \eqref{llm action}, $c_K$ facilitates in-context learning, i.e., learning via demonstration with examples, which can be explained through the lens of a recent theory on LLM latent spaces \cite{Jiang:2023}. Precisely, consider an $\epsilon$-ambiguity message $x$, defined as a message for which $p(\phi | x) \ge 1- \epsilon(x)$ The term $\epsilon(x)\in [0,1)$ represents the degree of ambiguity, and $p(\phi | x)$ is the possibility of inferring the original intention $\phi$ from $x$. According to Proposition 3 in \cite{Jiang:2023}, the difference between LLM's prediction probability $p(m^{\text{dn},t}|\tilde{m}^{\text{up},t}, c_K, i_{K+1})$ using $K$-shot in-context learning samples $c_K$ and the true conditional probability $q(m^{\text{dn},t}|i_{K+1}, \phi_{c_K})$ with an unknown common intention $\phi_{c_K}$ is bounded as follows:
\begin{equation}\tag{6}
\begin{aligned}
|p(m^{\text{dn},t}|\tilde{m}^{\text{up},t}, c_K, i_{K+1}) - q(m^{\text{dn},t}|i_{K+1}, \phi_{c_K})| \\
\le \epsilon(\tilde{m}^{\text{up},t}) \epsilon(i_{K+1}) \prod_{k=1}^{K} \epsilon(i_k, o_k).
\end{aligned}
\end{equation}
Since the rightmost term decreases with $K$ following a power law, only a few $K$ examples can significantly enhance the LLM's prediction accuracy through in-context learning.

Fig.~\ref{LSC_prompts} displays examples of the received UL messages $\tilde{m}^{\text{up},t}$ and the produced DL messages $m^{\text{dn},t}$ at the BS. Low SNR distorts the UL messages as observed by comparing Fig.~\ref{LSC_prompts}a and \ref{LSC_prompts}b. Due to the UL message distortion, the DL message in Fig.~\ref{LSC_prompts}d yields wrong actions $m_1^{a,t},$ and $m_2^{a,t}$ and inappropriate rationale $m^{\text{res},t}$, in contrast to proper actions and rationale at higher SNR in Fig.~\ref{LSC_prompts}c. Finally, Fig.~\ref{LSC_prompts}d shows the DL message without in-context learning, i.e., $K=0$, which fails to understand the given task, underscoring the importance of the few-shot examples $c_2$ in Fig.~\ref{LSC_prompts}c and \ref{LSC_prompts}d.

\section{Language-Guided Emergent Communication}

\begin{table}[t]
\caption{Computing cost comparison among EC, LSC, and LEC (using \cite{Wang:2023} for UEs in LSC and `calflops' \cite{calflops} library for the rest).}
\begin{center}
\begin{tabular}{ccccc}
\toprule
  &  \multicolumn{2}{c}{\textbf{FLOPs}} & \multicolumn{2}{c}{\textbf{Number of Parameters}}   \\ 
  \cline{2-5}
  &  UE & BS & UE & BS   \\ 
  \hline
\textbf{LSC}  & 22.5G & 17.57T & 109M & 68.71B \\ 
\textbf{EC}   & 25.63M & 34.05K & 278.31K & 2.24K   \\ 
\textbf{LEC}   & 25.63M & 34.05K & 278.31K & 2.24K \\
\bottomrule
\end{tabular}
\label{Model size comparison}
\end{center}
\end{table}

As studied in Sec.~\ref{Sec:ECvsLSC}, in-context learning enables control based on LSC without updating the LLM model, as opposed to EC necessitating MADRL training. However, LSC entails significant computing costs per inference due to the large model sizes. As shown in Table \ref{Model size comparison}, the floating point operations per second (FLOPS) for LSC at each UE is $890$x greater than that of EC when using the BLIP \cite{Li:2022} as $G_{\text{gen},j}(\cdot)$. Furthermore, the BS equipped with the LlaMA2-70B model \cite{Touvron:2023} as $G_{\text{LLM}}(\cdot)$ incurs FLOPS that is $520$ million times larger than that of EC. To mitigate this, we propose LEC that transfers the LSC's knowledge into EC via KD as elaborated next.



\subsection{Teacher Knowledge Construction via LSC}

We construct teacher knowledge by recording the top-$L$ outputs $H^{\star}$ from the set of $N$ episodes $H$, generated by LSC.

\begin{enumerate}
    \item \textbf{$N$-Episode Trajectory Generation via LSC}:    
    LSC outputs $H$ is a set defined as $H = \{ H^1,H^2, \\ \ldots, H^N\}$. The $n$-th episode $H^{n}$ consists of the records of all the UEs $\{H_1^{n}, H_2^{n}, \ldots H_{|\mathcal{J}|}^{n}\}$, where 
    \begin{align}\tag{7}
    H_j^{n} \!\!=\!\! \bigg\{(\mathbf{z}_{j}^{1,n}\!\!, a_{j}^{1,n}\!), (\mathbf{z}_{j}^{2,n}\!\!, a_{j}^{2,n}\!), \cdots, (\mathbf{z}_{j}^{T_j^{n},n}\!\!, a_{j}^{T_j^{n},n}\!) \bigg\}.
    \end{align}
    
    \item \textbf{Invalid Action and Circular Path Removal}: 
    Elements in $H_j^{n}$ with  $a_{j}^{t,n} = (0,0)$  indicate invalid action, which should be removed. In addition, circular paths whereby a return to a prior location within the same episode are removed as follows. If there exist $t_1$ and $t_2$ that satisfy $t_1 < t_2$ and  $\mathbf{z}_{j}^{t_1,n} = \mathbf{z}_{j}^{t_2 ,n}$, eliminate $(\mathbf{z}_{j}^{t,n}, a_{j}^{t,n})$ pairs where $t_1 \le t < t_2$. Retention of such ineffective pairs could misguide EC into learning wrong behaviors. 
    
    \item \textbf{Top-$L$ Trajectory Selection}:
    $L$ trajectories are chosen as teacher knowledge among refined $H$ by finding $n_\ell$:
    
\begin{align}\tag{8}
n_\ell = \underset{n\in \mathcal{N} \backslash{\mathcal{N}^\star_{\ell-1}} }{\arg\min} \left\{\max_{j} T_j^n + \sum_{j=1}^{|\mathcal{J}|}\sum_{t=1}^{T_j} \frac{P_r}{|h_{\mathbf{z}_j^{t,n}}|^2} \right\}
\end{align}
for $\ell = 1,2,\ldots, L$, where $\mathcal{N} = \{1, 2, \cdots, N\}$, $\mathcal{N}^\star_{\ell} = \{ n_1, n_2, \cdots, n_{\ell}\} $, and $\mathcal{N}^\star_0= \phi$. The process finds $L$ outputs within the minimum sum of the maximum episode length across all UEs, denoted by $\max_{j} T_j$, and UL power consumption $P_r / |h_{\mathbf{z}_j^{t,n}}|^2$. Consequently, the teacher knowledge $H^{\star}$ is constructed as $H^{\star} = \{H^{n_1}, H^{n_2}, \ldots, H^{n_L} \}$.
\end{enumerate}

\begin{figure}[t]
  \centering
  \begin{subfigure}{.24\columnwidth}
    \includegraphics[width=\linewidth]{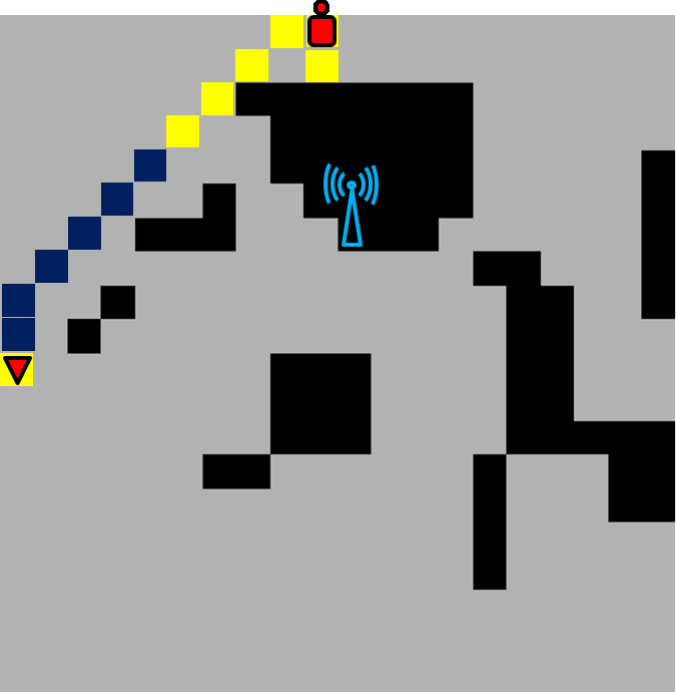}
    \caption{EC.}
    \label{EC_traj_2}
  \end{subfigure}\hfill
  \begin{subfigure}{.24\columnwidth}
    \includegraphics[width=\linewidth]{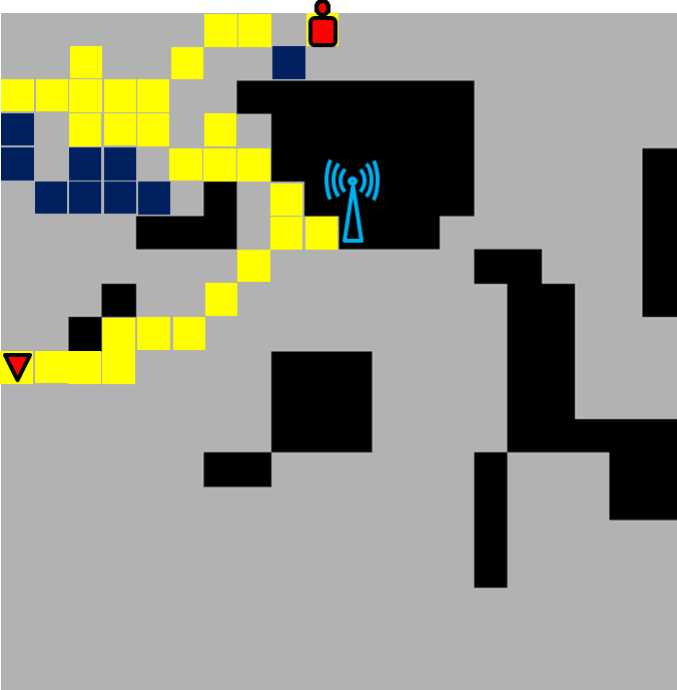}
    \caption{LSC.}
    \label{LSC_traj_2}
  \end{subfigure}\hfill
  \begin{subfigure}{.24\columnwidth}
    \includegraphics[width=\linewidth]{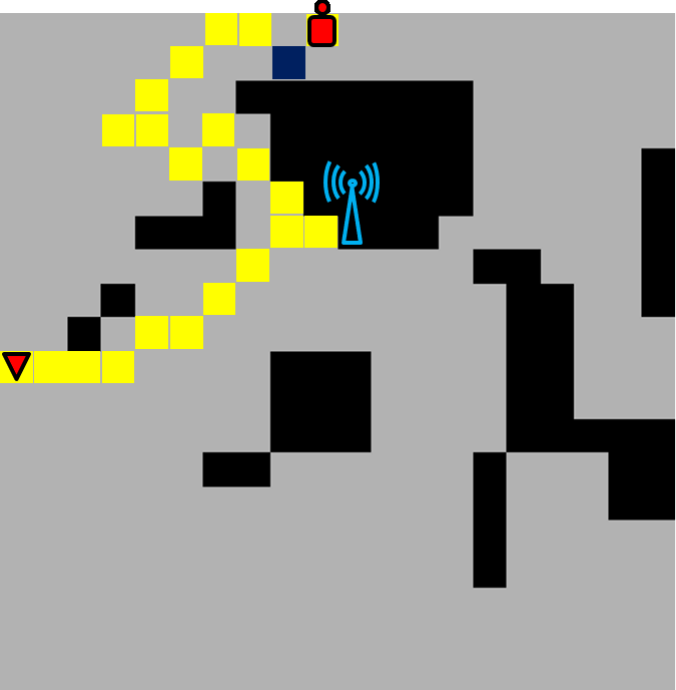}
    \caption{Refined~LSC.}
    \label{teacher_traj_2}
  \end{subfigure}\hfill
  \begin{subfigure}{.24\columnwidth}
    \includegraphics[width=\linewidth]{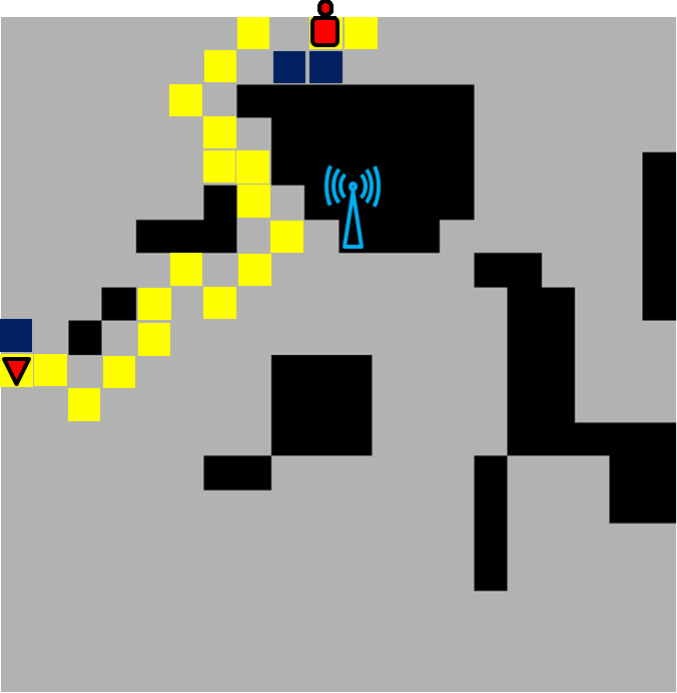}
    \caption{LEC.}
    \label{LEC_traj_2}
  \end{subfigure}
  \vspace{0.3cm}
  \caption{Trajectory comparison among different schemes : EC, LSC, refined LSC to utilize as teacher output for LEC, and LEC.}
  \label{trajectories}
\end{figure}

\subsection{Knowledge Distillation from LSC to EC}
Based on the constructed teacher knowledge involving LSC's top-$L$ refined trajectories, LEC applies KD by penalizing the difference between the outputs of EC and the teacher as elaborated next.

\begin{enumerate}
\item  \textbf{KLD Regularization}:
A Kullback-Leibler divergence (KLD) term $\mathcal{L}_\text{KLD} := D_{\text{KLD}}\left(p(a_j^t|\theta) || p(a_j^t|H^\star)\right)$ is incorporated into the loss function when the UE's current position $\mathbf{z}_j^t$ aligns with $\mathbf{z}_{j}^{\ell}$ to let LEC mimic the action of $H^{\star}$. The teacher knowledge $H^{\star}$ offers possible actions with discrete probability distribution $p(a_j^t|H^{\star})$ calculated as
\begin{align}\tag{9}
p(a_j^t | H^\star\!) \!=\! 
\begin{cases}
\frac{{\sum_{n=n_1}^{n_L} \!\sum_{u=1}^{T_j^n} \!\mathbf{1}_{\mathcal{B}}(u,\!n)}}{{\sum_{n=n_1}^{n_L}\! \sum_{u=1}^{T_j^n} \!\mathbf{1}_{\mathcal{D}}(u,\!n)}}, & \text{if }\mathcal{D} \ne \emptyset \\
0, & \text{otherwise.}
\end{cases}
\label{teacher action pdf}
\end{align}
Function $\mathbf{1}_{\mathcal{D}}(u,n)$ returns 1 if $(u,n) \in \mathcal{D}$ where  $\mathcal{D} := \{(u,n) \mid \mathbf{z}_j^t = \mathbf{z}_j^{u,n}\}$ and $\mathbf{1}_{\mathcal{B}}(u,n)$ returns 1 if $(u,n) \in \mathcal{B}$ where  $\mathcal{B} := \{(u,n) \mid 
\mathbf{z}_j^t = \mathbf{z}_j^{u,n}, a_j^t = a_j^{u,n}\}$.

\item \textbf{KLD Divergence Mitigation}: 
If $\mathcal{D} = \emptyset$, the KLD diverges, motivating UEs to replicate the teacher's path by providing rewards whenever a UE's position coincides with any $\mathbf{z}_{j}^{\ell}$. So, reward term of LEC $\tilde{r}_j^t$ is defined as:


\begin{equation}\tag{10}
    \tilde{r}_{j}^t = r_j^t + 0.1 \cdot \mathbf{1}_{\mathbf{z}_j^t = \mathbf{z}_{j}^{\ell}}.
    \label{reward function_lec}
\end{equation}
Adding the new reward term promotes adherence to the paths that are well-established in the teacher's model. 
\end{enumerate}

In summary, LEC finds an optimal parameter set $\theta_{\text{LEC}}^{\star}$ via solving (\ref{LEC loss function}), where $\lambda$ is a hyperparameter that adjusts the weight of the KLD term. Minimizing the KLD guides the UEs to the aggregated action preferences of the teacher model, thereby guiding them to locations previously navigated by the teacher model, leading to a higher reward. These processes facilitate the transfer of strategic navigation knowledge from LSC to EC, thus able to determine policy via 
\begin{align} \setcounter{equation}{10}
\text{(\textbf{LEC})}\quad a_j^t = \mathop{\mathrm{arg\,max}}_a p(a | s_j^t, h_j^t, \tilde{m}_j^{\text{dn},t} ; \theta_{\text{LEC}}^{\star}),
\end{align}
where $\theta_{\text{LEC}}^{\star}$ denotes the model after KD, given in~\eqref{LEC loss function}.
 
\begin{figure}[t]
  \centering
  \begin{subfigure}[b]{0.46\columnwidth}
    \includegraphics[width=\textwidth]{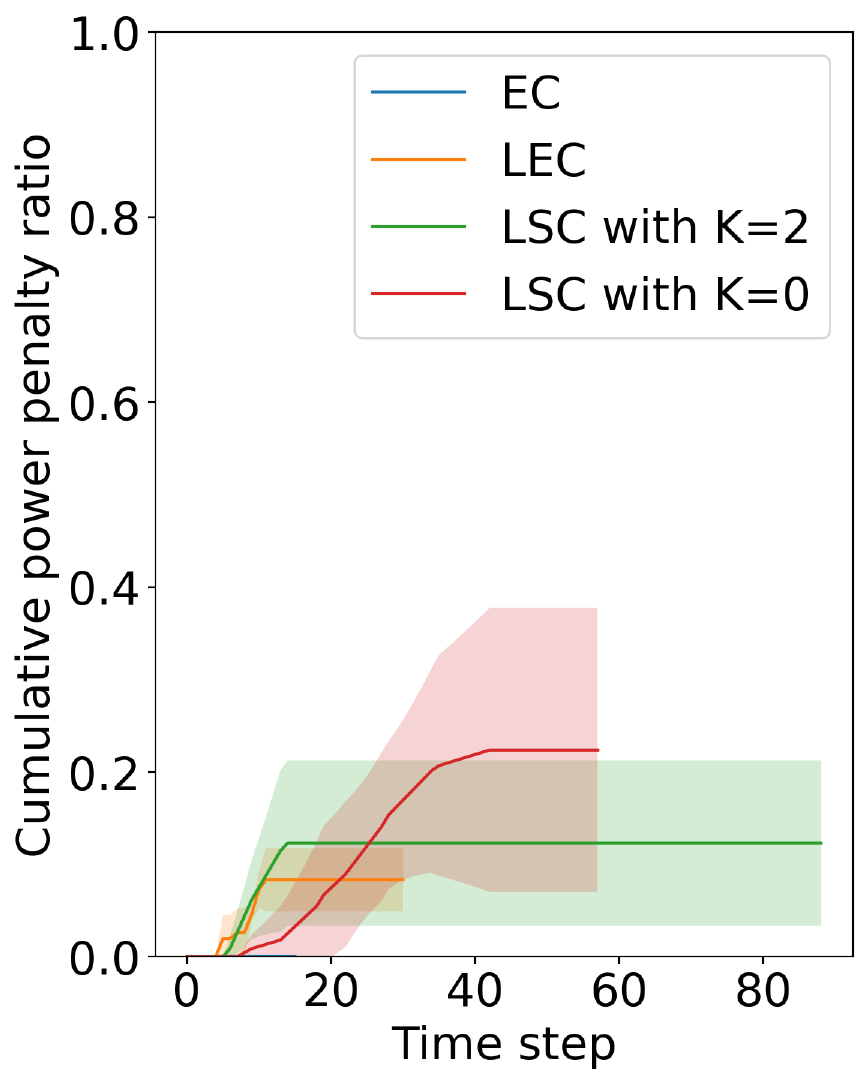}
    \caption{$\text{UE}_1$.}
    \label{UE1_CMA}
  \end{subfigure} \hspace{-0.1cm}
  \begin{subfigure}[b]{0.46\columnwidth}
    \includegraphics[width=\textwidth]{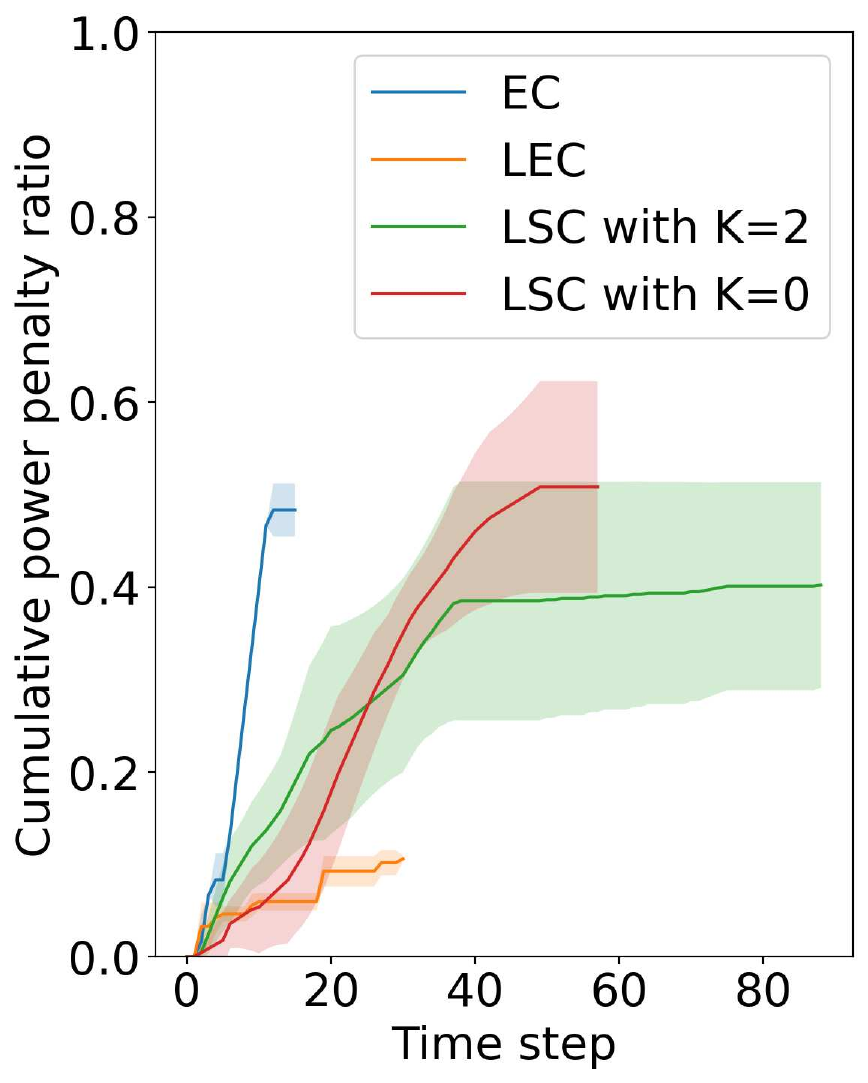}
    \caption{$\text{UE}_2$.}
    \label{UE2_CMA}
  \end{subfigure}
  \caption{Cumulative power penalty ratio (CPPR) of $\text{UE}_1$ and $\text{UE}_2$.}
  \label{CMA}
\end{figure}

\section{Simulation Results}

The environment depicted in Fig.~\ref{global map} and the channel coefficient at location $\mathbf{z}_j^t$, denoted by $h_{\mathbf{z}_j^t}$, are obtained using the ray-tracing simulator WiThRay from \cite{Choi:2023}. The environmental map has been discretized into a $20 \times 20$ grid, implying that both $W_x$ and $W_y$ are set to 20. The BS is located at the coordinates $(6,10)$. For the sake of simplification, the number of UEs is limited to two, i.e., $|\mathcal{J}|=2$, with their respective initial and target positions designated as $\mathbf{z}_{1}^{0} = (0,0)$, $\mathbf{z}_{2}^{0} = (0,9)$, $\mathbf{\hat{z}}_{1} = (9,9)$, and $\mathbf{\hat{z}}_{2} = (10,0)$. The carrier frequency and the bandwidth are set as $2.3$ GHz and $30.72$ MHz, respectively. 

Within LSC, the BLIP model is employed for the UEs. This model has been fine-tuned in advance to generate textual content enriched with action decision-relevant information. Moreover, the LlaMA2 GPTQ \cite{Frantar:2023}, a language model with 70 billion parameters, is integrated as $G_{\text{LLM}}(\cdot)$. Regarding EC, the following hyperparameters are a learning rate of $10^{-4}$, a total of $10,\!000$ training episodes, a discount factor $\gamma$ of 1, an initial exploration probability of $0.05$, an exploration probability decay ratio of $0.999$, and a complex emergent message length set to $8$. The experiments are conducted on NVIDIA Quadro RTX 8000 (4EA).

\begin{figure}[t]
\begin{center}
\includegraphics[width=.95\columnwidth]{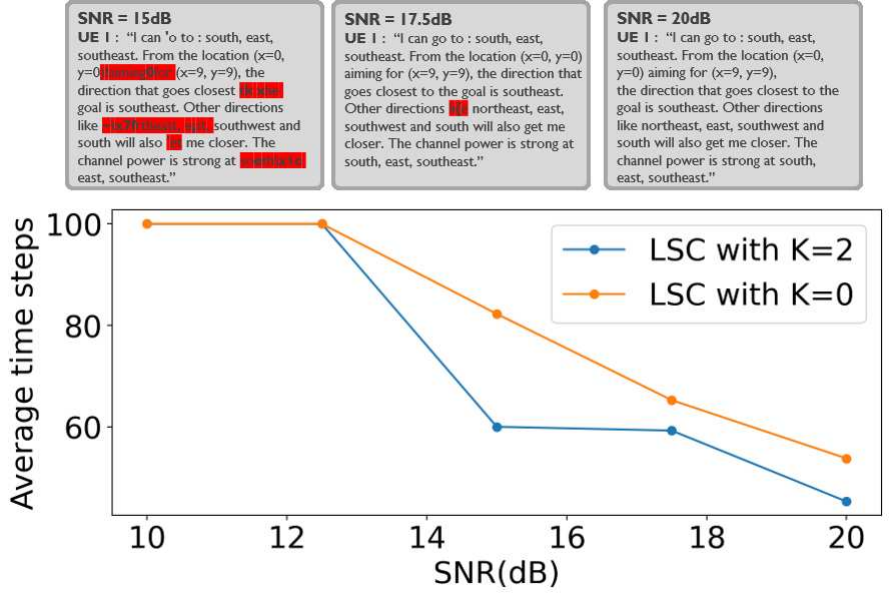}
\end{center}
\caption{Effect of in-context learning on distorted prompt.}
\label{in_context}
\end{figure}

\vspace{5pt}
\noindent\textbf{Multi-Modal Processing Capability and Path Uncertainty}:
Fig.~\ref{trajectories} shows $\text{UE}_2$ paths under EC, LSC, and LEC, with $\sqrt{\eta} = 3 \times 10^{-7}$, the global channel map illustrated as Fig.~\ref{global map}(c). In the grid, yellow and dark blue squares represent the UE paths, in which the locations are associated with $|h_{\mathbf{z}_j^t}|^2 \geq \eta  $ and $|h_{\mathbf{z}_j^t}|^2 < \eta$, respectively. EC only takes into account navigation maps, ignoring channel maps, which results in large proportion of dark blue regions in the trajectory. Thanks to considering both location and channel maps, LSC is better at avoiding dark blue regions. However, it often yields inefficient trajectories due to the LLM's enforcing inference despite high uncertainty, i.e., hallucinations, and the resultant errors propagating through $m_{t-1}^\text{bs}$ or its context window. To address this, LEC utilizes a refined trajectory in Fig.~\ref{trajectories}(c) from the original LSC's trajectory in Fig.~\ref{trajectories}(b). By selecting LSC's five trajectories out of 50 episodes, i.e., $N = 50$, $L=5$, the MADRL training is guided using $H^{\star}$ with $\lambda = 1$, resulting in LEC's efficient trajectories as shown in Fig.~\ref{trajectories}(d).

\vspace{5pt}
\noindent\textbf{Transmit Power Penalty and Travel Time}:
Fig.~\ref{CMA} examines the average cumulative power penalty ratio (CPPR) for each UE.  $\text{CPPR}_j^t$ is the ratio of the cumulative power penalty the $j$-th UE incurred up to a certain $t$ when all UEs have reached their destination, calculated as $\text{CPPR}_j^t = \sum_{u=1}^{t} \mathbf{1}_{\mathcal{P}}(u) /\max_j (T_j)$.
The function $\mathbf{1}_{\mathcal{P}} (u)$ returns $1$ if $u \in \mathcal{P}$ such that $\mathcal{P} := \{w \mid |h_{\mathbf{z}_j^w}|^2 < \eta \}$, and otherwise we have 0. The solid line represents the average CPPR of the eight least time-consuming scenarios out of twenty to compare among the cases that all UEs have arrived at the destinations, while the shaded areas denote the standard deviation from the mean. In Fig.~\ref{CMA}(a), the task for $\text{UE}_1$ is relatively easy thanks to its initial location and the destination, yielding low CPPRs for all schemes, only excluding LSC without in-context learning, i.e., $K=0$. Therefore, we hereafter focus only on $\text{UE}_2$ depicted in Fig.~\ref{CMA}(b). For $\text{UE}_2$, EC and LSC lead to high average CPPRs due to the incapability of processing multi-modal data and high path uncertainties, respectively. By contrast, LEC seeks for achieving LSC's top-$L$ trajectories, thereby yielding fast travel time with the lowest CPPR.


\begin{figure}[t]
\begin{center}
\includegraphics[width=.9\columnwidth]{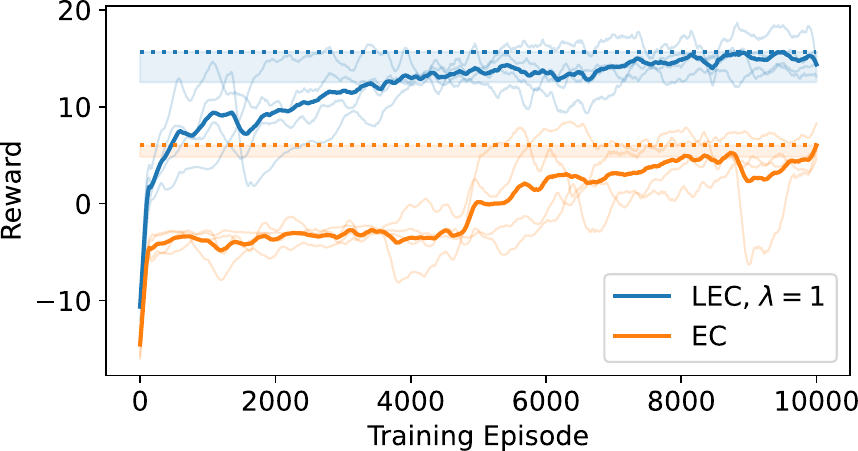}
\end{center}
\caption{Training performance and convergence of LEC and EC.}
\label{convergence}
\end{figure}


\vspace{5pt}
\noindent\textbf{Impact of In-Context Learning}:
Fig.~\ref{in_context} presents the average time steps of the two shortest and the two longest cases with LSC to reach the goal among 10 episodes with different values of SNR. We set quantized two-level global channel map as Fig.~\ref{global map}(d) to provide an option for $\text{UE}_2$ to find path with less detour. At low SNR, $\tilde{m}^{\text{up},t}$ is distorted, rendering LLM unable to comprehend the input prompt and subsequently decide a proper action. This results in a random selection of actions, analogous to EC's tendency to prioritize exploration over exploitation every time step. We demonstrate the impact of few-shot learning on ambiguity reduction when input prompts are distorted by comparing scenarios with $K=0$ and $K=2$. The results indicate that incorporating more few-shot examples significantly reduces ambiguity and leads to fewer wrong decisions when prompts are partially distorted like demonstrated at the figure.

\vspace{5pt}
\noindent\textbf{Convergence Acceleration via KD}:
Fig.~\ref{convergence} shows that LEC has accelerated convergence compared to EC. The thick lines are the average reward of EC and LEC along the training episodes, averaged over four different runs that are shown with transparent lines. For better visibility, the rewards are smoothed with Savitzky-Golay filter with polynomial order of $3$ and window length of $301$ episodes. To compare the convergence, we draw the convergence region, which is the range between the maximum reward and $80\%$ of the maximum reward with shaded areas within the graph. LEC’s average reward curve shows fast and stable convergence by staying within the shaded convergence region from episode $3,\!636$, whereas EC’s curve fails to stay within the shaded convergence region until episode $9,\!520$.

\section{Conclusion}
In this paper, we explored the effectiveness of an MADRL-based EC and an LLM-based LSC in a multi-agent, multi-modal remote control and navigation task. To reap the advantages of both EC and LSC, we proposed LEC by distilling the LSC's top-$L$ trajectory knowledge into EC, thereby achieving low computing costs at both training and inference with enhanced task performance. As a preliminary study, the current simulations rely on additive noise channels with two UEs. To enhance the feasibility of LEC, future research needs to incorporate realistic channels with more UEs. Furthermore, to cope with low SNR, DeepJSCC could be additionally utilized for the UL and DL message communication in LEC, which could be an interesting topic for future work.


\section*{Acknowledgement}
This research was supported by the MSIT(Ministry of Science and ICT), Korea, under the ITRC(Information Technology Research Center) support program(IITP-2024-2020-0-01787) supervised by the IITP(Institute of Information \& Communications Technology Planning \& Evaluation)
\bibliographystyle{IEEEtran}
\bibliography{references}

\end{document}